\documentclass[11pt]{article}
\usepackage{jair}
\usepackage{theapa}
\usepackage{tabularx}
\usepackage{xparse}
\usepackage{amsmath}
\usepackage{amsthm}
\usepackage{amssymb}
\usepackage{xspace}
\usepackage{relsize}
\usepackage{graphicx}
\usepackage{multirow}
\usepackage{url}
\usepackage{comment}

\usepackage[frozencache]{minted}

\usepackage{xr}

\newcommand{\subsubparagraph}[1]{}

\makeatletter
\let\@myref\ref

\newcommand{\refsec}[1]{Sec.\,\@myref{#1}}
\newcommand{\refseq}[1]{Sec.\,\@myref{#1}}
\newcommand{\refig}[1]{Fig.\,\@myref{#1}}
\newcommand{\refigs}[2]{Fig.\,\@myref{#1}-\@myref{#2}}

\newcommand{\reftbl}[1]{Table \@myref{#1}}
\newcommand{\refstep}[1]{Step \@myref{#1}}
\newcommand{\refalgo}[1]{Algorithm \@myref{#1}}
\newcommand{\refchap}[1]{Chapter \@myref{#1}}
\newcommand{\reflst}[1]{List \@myref{#1}}
\newcommand{\refeq}[1]{Eq. \@myref{#1}}

\makeatother

\newcounter{list}[section]

\newcommand{\braces}[1]{{\left\{#1\right\}}}

\newcommand{\defun}[1]{%
\makeatletter
\expandafter\def\csname the#1\endcsname{\text{\it #1}}
\expandafter\def\csname #1\endcsname ##1{\csname the#1\endcsname\left(##1\right)}%
\makeatother
}

\newcommand{\defsetop}[2]{%
\makeatletter
 \expandafter\def\csname #1\endcsname ##1##2##3{%
  \expandafter\def\csname #1arg\endcsname{##1}%
  \expandafter\def\csname #1set\endcsname{##2}%
  \expandafter\def\csname #1cond\endcsname{##3}%
  \braces{##1##2\mid #2 ##3}%
 }%
\makeatother%
}

\defun{precond}
\defun{condition}
\defun{params}
\defun{type}
\defun{proc}
\defun{cost}

\defun{push}
\defun{pref}

\defun{init}
\defun{goal}
\defun{objects}
\defun{task}

\defun{parent}
\defun{plateau}

\defsetop{filter}{}
\defsetop{map}{}

\newcommand{\pddl}[1]{\textsf{\small #1}}

\makeatletter
\NewDocumentCommand{\todo}{s O{} m}{%
  \IfBooleanTF#1%
    {\@todos{#2}{#3}}%
    {\@todons{#2}{#3}}}
\newcommand{\@todons}[2]{}
\newcommand{\@todos}[2]{}
\makeatother

\def\_{\\[-0.3em]}

\makeatletter

\newcommand{\newheuristic}[2]{%
 \def#1{%
  \ifmmode%
  h^\text{#2}\xspace%
  \else%
  \text{#2}\xspace%
  \fi%
 }%
}

\newheuristic{\lmcut}{LMcut}
\newheuristic{\mands}{M\&S}
\newheuristic{\pdb}{PDB}
\newheuristic{\ff}{FF}
\newheuristic{\ce}{CEA}
\newheuristic{\cg}{CG}
\newheuristic{\ad}{add}
\newheuristic{\lc}{LC}

\newcommand{\newUnitCostHeuristic}[2]{%
 \def#1{%
  \ifmmode%
  \hat{h}^\text{#2}\xspace%
  \else%
  \text{#2}\xspace%
  \fi%
 }%
}

\newUnitCostHeuristic{\lmcuto}{LMcut}
\newUnitCostHeuristic{\mandso}{M\&S}
\newUnitCostHeuristic{\ffo}{FF}
\newUnitCostHeuristic{\ceo}{CEA}
\newUnitCostHeuristic{\cgo}{CG}
\newUnitCostHeuristic{\ado}{add}
\newUnitCostHeuristic{\gco}{GoalCount}
\newUnitCostHeuristic{\lco}{LC}

\makeatother

\newcommand{\before}{pre}
\newcommand{\after}{suc}

\def\ref{\todo{Do not use ``ref'' directly!}}

\jairheading{1}{1993}{1-15}{6/91}{9/91}
\ShortHeadings{Photo-Realistic Blocksworld Dataset}{Asai}
\firstpageno{1}

\author{
\name Masataro Asai \email Masataro.Asai\textcircled{$\alpha$}ibm.com\\
\addr IBM Research}
\title{Photo-Realistic Blocksworld Dataset}

\begin{document}
\maketitle

\begin{abstract}
In this report, we introduce an artificial dataset generator for Photo-realistic Blocksworld domain. Blocksworld is one of the oldest high-level task planning domain that is well defined but contains sufficient complexity, e.g., the conflicting subgoals and the decomposability into subproblems. We aim to make this dataset a benchmark for Neural-Symbolic integrated systems and accelerate the research in this area. The key advantage of such systems is the ability to obtain a symbolic model from the real-world input and perform a fast, systematic, complete algorithm for symbolic reasoning, without any supervision and the reward signal from the environment.
\end{abstract}

\section{Introduction}


Blocksworld is one of the earliest symbolic AI planning domains that
have been traditionally expressed in a standardized PDDL language \cite{McDermott00}.
With an advent of modern machine learning systems that are capable of automatically deriving
the symbolic, propositional representation of the environment from the raw inputs \cite{Asai2018,kurutach2018learning},
it became realistic to directly solve the visualized version of the classical planning problems like Blocksworld.

To accelerate the research in this area, we publish Photo-realistic
Blocksworld Dataset Generator at
\url{https://github.com/ibm/photorealistic-blocksworld},
a system that renders the realistic Blocksworld images and put the results in a compact Numpy archive.

\begin{figure}[htbp]
 \centering
 \includegraphics[width=0.49\linewidth]{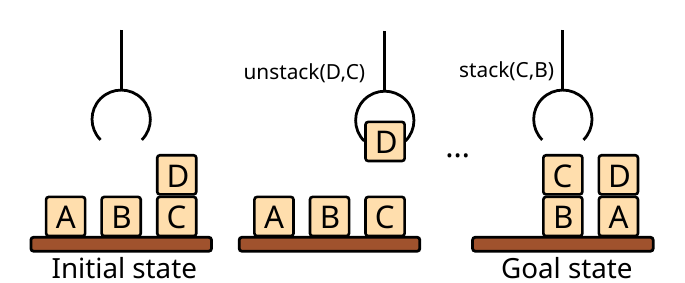}
 \includegraphics[width=0.49\linewidth]{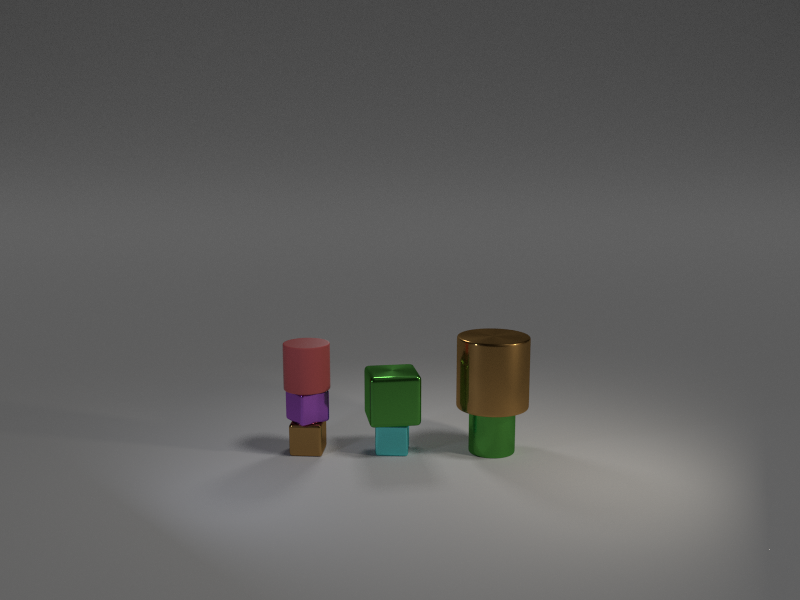}
 \caption{
(\textbf{Left}): A simplified illustration of Blocksworld domain.
(\textbf{Right}): An example photo-realistic rendering of a Blocksworld state in this dataset.
}
\label{blocks}
\end{figure}

%

In this document, we first describe the Blocksworld domain, then proceed
to the details of the dataset and the extension added to the
environment.  Finally, we show the results of some baseline experiments
of an existing Neural-Symbolic planning system Latplan \cite{Asai2018} applied to this dataset.

\section{The Blocksworld Domain}

Blocksworld is a domain which was proposed in SHRDLU \cite{winograd1971procedures}
interactive natural language understanding program, which takes a command input given as a natural language text
string, invokes a planner to achieve the task, then display the result.
It is still used as a testbed for evaluating various computer vision / dialog systems \cite{perera2018building,bisk2017learning,she2014back,johnson2009joint}
Most notably, SHRDLU heavily inspires the CLEVR standard benchmark dataset for Visual Question Answering \cite{johnson2017clevr}.

Blocksworld domain typically consists of several stacks of wooden blocks
placed on a table, and the task is to reorder and stack the blocks in a
specified configuration.
The blocks are named (typically assigned an alphabet) and are supposed to be identifiable,
but in the real world, the color could replace the ID for each block.

In order to move a block $b$, the block should be clear of anything on
the block itself, i.e., it should be on top of a stack, so that the arm
can directly grasp a single block without hitting another object. This
restriction allows the arm to grasp only a single block at a time.
While this condition could be expressed in a quantified first-order
formula using a predicate \pddl{on}, e.g., $\forall b_2; \lnot
\pddl{on}(b,b_2)$, typically this is modeled with an auxiliary predicate
\pddl{clear} with an equivalent meaning.

There are a couple of variations to this domain such as the one which explicitly involves the concept of a robotic arm.
In this variation, moving a block is decomposed into two actions, e.g., \pddl{pick-up} and \pddl{put-down},
where \pddl{pick-up} action requires a \pddl{handempty} nullary predicate to be true.
The \pddl{handempty} predicate indicates that the arm is currently holding nothing, and can grasp a new block.
These predicates could be trivially extended to a multi-armed scenario where \pddl{pick-up} and \pddl{handempty} could take an additional
\pddl{?arm} argument, but we do not add such an extension in our dataset.

When modeled as a pure STRIPS domain where the disjunctive preconditions (\textit{or}'s in the preconditions) are
prohibited, the actions can be further divided into \pddl{pick-up/put-down} actions and \pddl{unstack/stack} action,
wherein \pddl{pick-up/put-down} the arm grasps/releases a block from/onto the table, while
in \pddl{stack/unstack} the arm grasps a block from/onto another block.
An example PDDL representation of such a model is presented in \refig{4-op-blocks}.

The domain has further variations such as those containing the size
constraint (a larger block cannot be put on a smaller block) or
other non-stackable shapes (spheres or pyramids)
\cite{gupta1992complexity}, but we do not consider them in this dataset.

\begin{figure}[htbp]
\begin{minted}{common-lisp}
(define (domain blocks)
  (:requirements :strips)
  (:predicates (on ?x ?y)
	       (ontable ?x)
	       (clear ?x)
	       (handempty)
	       (holding ?x))

  (:action pick-up
	     :parameters (?x)
	     :precondition (and (clear ?x) (ontable ?x) (handempty))
	     :effect
	     (and (not (ontable ?x))
		   (not (clear ?x))
		   (not (handempty))
		   (holding ?x)))

  (:action put-down
	     :parameters (?x)
	     :precondition (holding ?x)
	     :effect
	     (and (not (holding ?x))
		   (clear ?x)
		   (handempty)
		   (ontable ?x)))
  (:action stack
	     :parameters (?x ?y)
	     :precondition (and (holding ?x) (clear ?y))
	     :effect
	     (and (not (holding ?x))
		   (not (clear ?y))
		   (clear ?x)
		   (handempty)
		   (on ?x ?y)))
  (:action unstack
	     :parameters (?x ?y)
	     :precondition (and (on ?x ?y) (clear ?x) (handempty))
	     :effect
	     (and (holding ?x)
		   (clear ?y)
		   (not (clear ?x))
		   (not (handempty))
             (not (on ?x ?y)))))
\end{minted}
\caption{4-op Blocksworld in PDDL.}
\label{4-op-blocks}
\end{figure}



The problem has been known for exhibiting the subgoal interactions
caused by the delete effects, which introduced the well-known Sussman's
anomaly
\cite{sussman1973computational,sacerdoti1975nonlinear,waldinger1975achieving,mcdermott1985introduction,norvig1992paradigms}.
The anomaly states that achieving some subgoal requires destroying an already achieved subgoal once (and restore it later).
This characteristic makes the problem difficult as the
agent needs to consider how to construct the subgoals in the correct order (subgoal ordering).
Later, the class of problem containing such interaction was shown to make the planning problem PSPACE-complete
while the delete-free planning problem was shown to be only NP-complete \cite{backstrom1995complexity}.

Solving the Blocksworld problem suboptimally, i.e.,
without requiring to return the fewest number of moves that solves the problem, is tractable.
Indeed, we can solve the problem in linear time in the following algorithm:
First, \pddl{put-down} all blocks onto the floor, then construct the desired stacks.
This procedure finishes in a number of steps linear to the number of blocks.
In contrast, the decision problem version of the optimization problem, i.e.,
``Given an integer $L>0$, is there any path that achieves the goal state under $L$ steps?''
is shown to be NP-hard \cite{gupta1992complexity}.

More recently, an in-depth analysis on the suboptimal / approximation
algorithms and the problem generation method was performed on the
domain \cite{slaney2001blocks}, showing that the domain is still able
to convey valuable lessons when analyzing the performance of
modern planners.

\begin{figure}[htbp]
 \centering
 \includegraphics[width=0.5\linewidth]{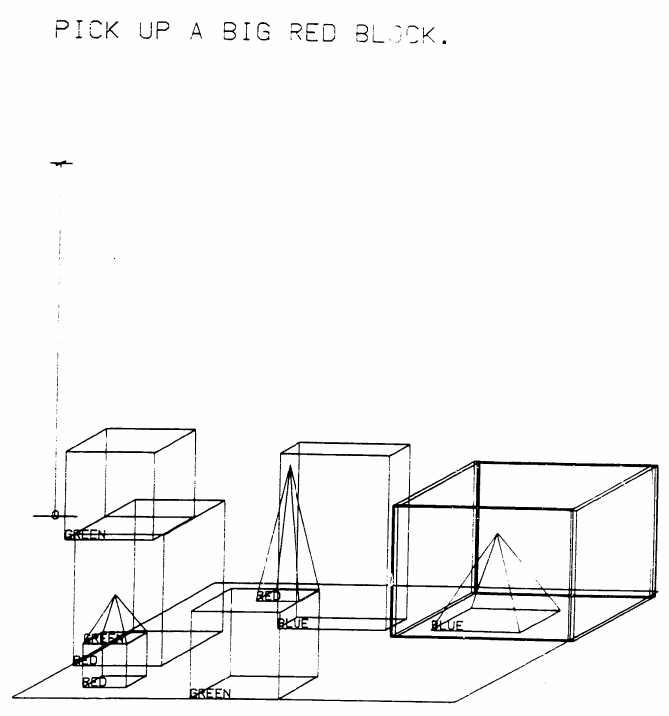}
 \caption{A demonstration of SHRDLU natural language understanding computer program. The image was taken from Winograd's original paper \cite{winograd1971procedures}.}
 \label{shrdlu}
\end{figure}

\section{The Dataset}

The code base for the dataset generator is a fork of CLEVR dataset
generator where the rendering code was reused, with modifications to the
logic for placing objects and enumerating the valid transitions.
It renders the constructed scenes in Blender 3D rendering engine which
can produce a photorealistic image generated by ray-tracing (\refig{blocks}, right).

In the environment,
there are several cylinders or cubes of
various colors and sizes and two surface materials (Metal/Rubber)
stacked on the floor, just like in the usual STRIPS Blocksworld domain.
The original environment used in SHRDLU contains shapes other than cubes such as pyramids, which are not stackable.
While we do not use those unstackable shapes in the generators by default,
it can be supported with a few modifications to the code.

Unlike the original blocksworld, three actions can be performed in this environment:
\texttt{move} a block onto another stack or the floor,
and \texttt{polish}/\texttt{unpolish} a block, i.e., change the
surface of a block from Metal to Rubber or vice versa.
All actions are applicable only when the block is on top of a stack or on the floor.
The \texttt{polish}/\texttt{unpolish} actions allow changes in the non-coordinate features of the object vectors,
which adds additional complexity.

\begin{figure}[htbp]
 \centering
 \includegraphics[width=\linewidth]{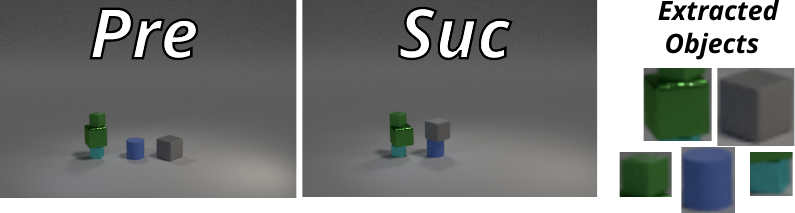}
 \caption{
An example Blocksworld transition.
Each state has a perturbation from the jitter in the light positions, object locations, object rotations, and the ray-tracing noise.
Objects have the different sizes, colors, shapes and surface materials.
Regions corresponding to each object in the environment are extracted
according to the bounding box information included in the dataset
generator output, but is ideally automatically extracted by object
recognition methods such as YOLO \cite{redmon2016you}.  Other objects
may intrude the extracted regions, as can be seen in the extraction of
the green cube (top-left) which also contains the bottom edge of the
smaller cube.  } \label{blocks-example}
\end{figure}

The dataset generator takes the number of blocks and the maximum number of stacks
that are allowed in the environment. No two blocks are allowed to be of the same color, and
no two blocks are allowed to have the same shape and size.

The generator produces a 300x200 RGB image and a state description which contains
the bounding boxes (bbox) of the objects.
Extracting these bboxes is an object recognition task we do not address in this paper,
and ideally, should be performed by a system like YOLO \cite{redmon2016you}.

The generator comes with a postprocessing program,
which extracts and resizes the image patches in the bboxes to 32x32 RGB.
It stores the image patches as well as the bbox vector $(x_1,y_1,x_2,y_2)$ in a single, compressed Numpy format (.npz)
which is easily loaded from a python program environment.
The agent that performs a Machine Learning task on this environment is allowed to take the original rendering
as well as this segmented dataset.

The generator enumerates all possible states/transitions
(480/2592 for 3 blocks and 3 stacks; 80640/518400 for 5 blocks and 3 stacks).
Since the rendering could take time, the script provides support for a distributed cluster environment
with a batch scheduling system.

\section{Baseline Performance Experiment}

To demonstrate the baseline performance of an existing Neural-Symbolic planning system,
We modified Latplan
\cite{Asai2018} image-based planner, a system that operates on a
discrete symbolic latent space of the real-valued inputs and runs Dijkstra's/A* search using a state-of-the-art
symbolic classical planning solver. We modified Latplan to take the set-of-object-feature-vector input rather than images.

Latplan system learns the binary latent space of an
arbitrary raw input (e.g. images) with a Gumbel-Softmax variational autoencoder (State AutoEncoder network, SAE),
learns a discrete state space from the transition examples, and runs a symbolic, systematic search
algorithm such as Dijkstra or A*  search which guarantee the optimality of the
solution.  Unlike RL-based planning systems, the search agent does not contain the learning
aspects. The discrete plan in the latent space is mapped back to the raw image
visualization of the plan execution, which requires the reconstruction capability of (V)AE.
A similar system replacing Gumbel Softmax VAE with Causal InfoGAN was later proposed \cite{kurutach2018learning}.

When the network learned the representation, it guarantees that the
planner finds a solution because the search algorithm being used
(e.g. Dijkstra) is a complete, systematic, symbolic search algorithm,
which guarantees to find a solution whenever it is reachable in the state space.
If the network cannot learn the
representation, the system cannot solve the problem and/or return the
human-comprehensive visualization.

We generated a dataset for a 4-blocks, 3-stacks environment, whose
search space consists of 5760 states and 34560 transitions.
We provided 2500 randomly selected states for training the autoencoder.

We tested the generated representation with AMA$_1$ PDDL generator \cite{Asai2018}
and the Fast Downward \cite{Helmert04} classical planner.
AMA$_1$ is an oracular method
that takes the entire raw state transitions, encode each $\braces{\before_i, \after_i}$ pair with the SAE,
then instantiate each encoded pair into a grounded action schema.
It models the ground truth of the transition rules,
thus is useful for verifying the state representation.
Planning fails when SAE fails to encode a given init/goal image into
a propositional state that exactly matches one of the search nodes.
%
While there are several learning-based AMA methods that approximate AMA$_1$
(e.g. AMA$_2$ \cite{Asai2018} and Action Learner \cite{amado2018goal,amado2018lstm}),
there is information loss between the learned action model and the original search space generated.

We invoked Fast Downward with blind heuristics in order to remove its effect.
This is because AMA$_1$ generates a huge PDDL model containing all transitions
which result in an excessive runtime for initializing any sophisticated heuristics.
The scalability issue caused by using a blind heuristics is not an issue
since the focus of this evaluation is on the feasibility of the representation.

We solved 30 planning instances generated by
taking a random initial state and choosing a goal state by the 3, 7, or 14 steps random walks (10 instances each).
The result plans are inspected manually and checked for correctness.
While Latplan returned plans for all 30 instances, the plans were
correct only in 14 instances due to the error in the reconstruction (Details in
\reftbl{blocksworld-planning-results}). Apparently, the training seems more difficult
than the domains tested in \cite{Asai2018} due to the wider variety of disturbances in the environment.
This could be further improved
by analyzing and addressing the various deficiency of the latent representation
learned by the network, and by the hyperparameter tuning for the better accuracy.

\begin{table}[htbp]
 \centering
 \begin{tabular}{|c|c|}
  Random walk steps & The number of solved instances \\
  used for generating & (out of 10 instances each) \\
  the problem instances &   \\\hline
  3  & 7 \\
  7  & 5 \\
  14 & 2 \\\hline
 \end{tabular}
 \caption{The number of instances solved by Latplan using a VAE.}
 \label{blocksworld-planning-results}
\end{table}

\begin{figure}[htb]
 \centering
 \includegraphics[width=0.48\linewidth]{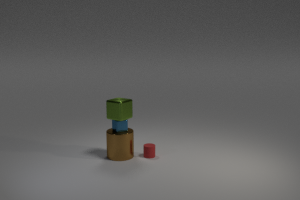}
 \includegraphics[width=0.48\linewidth]{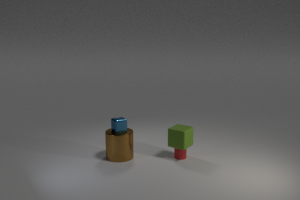}
 \caption{An example of a problem instance. (\textbf{Left}) The initial state. (\textbf{Right}) The goal state.
The planner should unpolish a green cube and move the blocks to the appropriate goal position, while also following the environment constraint that
the blocks can move or polished only when it is on top of a stack or on the floor.}
 \label{blocksworld-ig}
\end{figure}

\begin{figure}[htb]
 \includegraphics[width=0.24\linewidth]{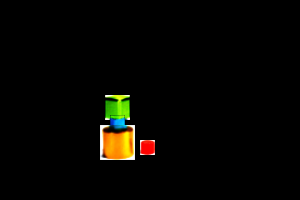}
 \includegraphics[width=0.24\linewidth]{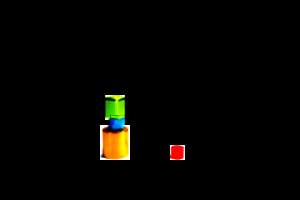}
 \includegraphics[width=0.24\linewidth]{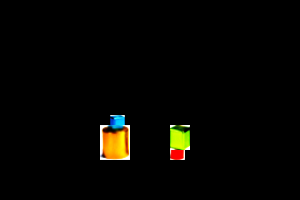}
 \includegraphics[width=0.24\linewidth]{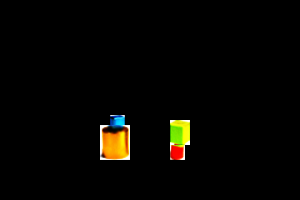}
 \caption{An example of a successful plan execution returned by Latplan.
Latplan found an optimal solution because of the underlying optimal search algorithm (Dijkstra).
}
 \label{blocksworld-success}
\end{figure}

\section{Related Work}

The crucial difference between this environment and Atari Learning Environment \cite{bellemare2013arcade} is twofold:
First, the action label is not readily available,
i.e., in ALE, the agent have the total knowledge on the possible combination of the key/lever inputs (up,down,left,right,fire),
while in our dataset the state transition pairs are not labeled by the actions.
The agent is required to find the set of actions by itself, possibly using an additional learning mechanism such as
AMA$_2$ system \cite{Asai2018}.

Second, in ALE, the scoring criteria / reinforcement signal is defined. In contrast, classical planning problems
like Blocksworld do not contain such signals except for the path length (also called an intrinsic reward in the RL field).
A possible extension of the baseline planning system presented above is to make it learn the goal conditions
rather than requiring the single goal state as the input.

There is another image dataset \cite{bisk2016towards} for an 3D environment that consists of blocks,
but the environment does not contain the combinatorial aspect
that is present in our dataset. Specifically, the environment does not contain the subgoal conflicts
as all blocks are initially located on the table.

\section{Discussion and Conclusion}

We introduced a generator for the Photo-Realistic Blocksworld dataset
and specified the environment that is depicted in it.
In the future, we aim to increase the variety of classical planning domains that are
expressed in the visual format, in a spirit similar to the Atari Learning Environment \cite{bellemare2013arcade},
or perhaps borrowing some data from it.

\fontsize{9.5pt}{10.5pt}
\selectfont

\end{document}